\documentclass{article}

\usepackage{microtype}
\usepackage{graphicx}
\usepackage{subfigure}
\usepackage{booktabs} 
\usepackage{amsmath}
\usepackage{amssymb}

\usepackage{hyperref}

\usepackage{listings}
\usepackage{xcolor}

\definecolor{codegreen}{rgb}{0,0.6,0}
\definecolor{codegray}{rgb}{0.5,0.5,0.5}
\definecolor{codepurple}{rgb}{0.58,0,0.82}
\definecolor{backcolour}{rgb}{0.95,0.95,0.92}

\lstdefinestyle{mystyle}{
    backgroundcolor=\color{backcolour},   
    commentstyle=\color{codegreen},
    keywordstyle=\color{magenta},
    numberstyle=\tiny\color{codegray},
    stringstyle=\color{codepurple},
    basicstyle=\sffamily\scriptsize,
    breakatwhitespace=false,         
    breaklines=true,                 
    captionpos=b,                    
    keepspaces=true,                 
    numbers=left,                    
    numbersep=5pt,                  
    showspaces=false,                
    showstringspaces=false,
    showtabs=false,                  
    tabsize=1
}

\usepackage{tikz}

\usepackage[accepted]{icml2020}

\icmltitlerunning{Learning Distributed Representations of Graphs with Geo2DR}

\begin{document}

\twocolumn[
\icmltitle{Learning Distributed Representations of Graphs with Geo2DR}




\begin{icmlauthorlist}
\icmlauthor{Paul Scherer}{cam}
\icmlauthor{Pietro Li\`o}{cam}
\end{icmlauthorlist}

\icmlaffiliation{cam}{Department of Computer Science and Technology, University of Cambridge, Cambridge, United Kingdom}

\icmlcorrespondingauthor{Paul Scherer}{paul.scherer@cl.cam.ac.uk}

\icmlkeywords{Machine Learning, ICML}

\vskip 0.3in
]



\printAffiliationsAndNotice{}  

\begin{abstract}
We present Geo2DR (\textit{Geo}metric to \textit{D}istributed \textit{R}epresentations), a GPU ready Python library for unsupervised learning on graph-structured data using discrete substructure patterns and neural language models. It contains efficient implementations of popular graph decomposition algorithms and neural language models in PyTorch which can be combined to learn representations of graphs using the distributive hypothesis. Furthermore, Geo2DR comes with general data processing and loading methods to bring substantial speed-up in the training of the neural language models. Through this we provide a modular set of tools and methods to quickly construct systems capable of learning distributed representations of graphs. This is useful for replication of existing methods, modification, or development of completely new methods. This paper serves to present the Geo2DR library and perform a comprehensive comparative analysis of existing methods re-implemented using Geo2DR across widely used graph classification benchmarks. Geo2DR displays a high reproducibility of results in published methods and interoperability with other libraries useful for distributive language modelling.

\end{abstract}

\section{Introduction}\label{intro}

Representation learning of graphs using neural networks has turned into a large and exciting hub of research driven by successive proposals of graph representation learning methods and datasets to apply them onto. A significant part of the activity has focused on \textit{Graph Convolutional Neural Networks} (GCNN). Such neural networks are characterised by \textit{graph convolutional} operators \cite{belkinspectral,chebnet,gcn} that serve as useful inductive biases for learning representations of nodes and other graph substructures. Gilmer et al. \yrcite{mpnn} generalised the convolution operator over irregular domains as a message passing scheme, allowing the specification of a full spectrum of methods as variants of this equation. Representations of entire graphs are then created through the successive application of message passing operations followed by different \textit{pooling} methods \cite{chebnet, diffpool, luzhnica} which aggregate node representations towards a single vector representation for the entire graph. 



The difficulty of reliably constructing GCNN models has driven the need for toolkits and libraries to facilitate their development for replication, extension and creation of new models. Several such libraries have been made including: \textit{Graph Nets} introduced by Battaglia et al. \yrcite{relationalinductivebias}, \textit{DGL} by Wang et al. \yrcite{wang2019dgl}, \textit{GEM} by Goyal et al. \yrcite{goyal}, and most recently \textit{PyTorch Geometric} by Fey and Lenssen \yrcite{Fey2019}. These libraries have greatly contributed to lowering the barrier of entry into GCNN research, fueling the development of novel methods and libraries supporting them in a healthy feedback cycle.

Alongside ongoing research into GCNNs and its variants, another approach has focused on extending graph kernel methods with neural language embedding methods \cite{deepgraphkernels, graph2vec, anonymouswalkembeddings} that exploit the distributive hypothesis to learn \textit{distributed representations} of graphs. This is a useful alternative inductive bias to model the vector space embeddings of graphs over the distribution of the discrete substructure patterns \textit{contextualising} them. Much like how the semantic meaning of words is similar to words that have similar context words around them \cite{harris}, distributed representations of graphs are inductively biased to be similar if they contain similar substructure patterns, and dissimilar otherwise. This perspective enables the construction of a powerful class of unsupervised representation learning methods.

However, to our knowledge, no toolkit currently exists for rapidly composing methods capable of learning distributed representations of graphs. This project, Geo2DR, aims to fill this gap. The library along with links to documentation, example methods, experiment replication, and supporting material can be found on the GitHub repository\footnote{https://github.com/paulmorio/geo2dr}.

\begin{figure*}
    \centering
    \includegraphics[width=0.8\textwidth]{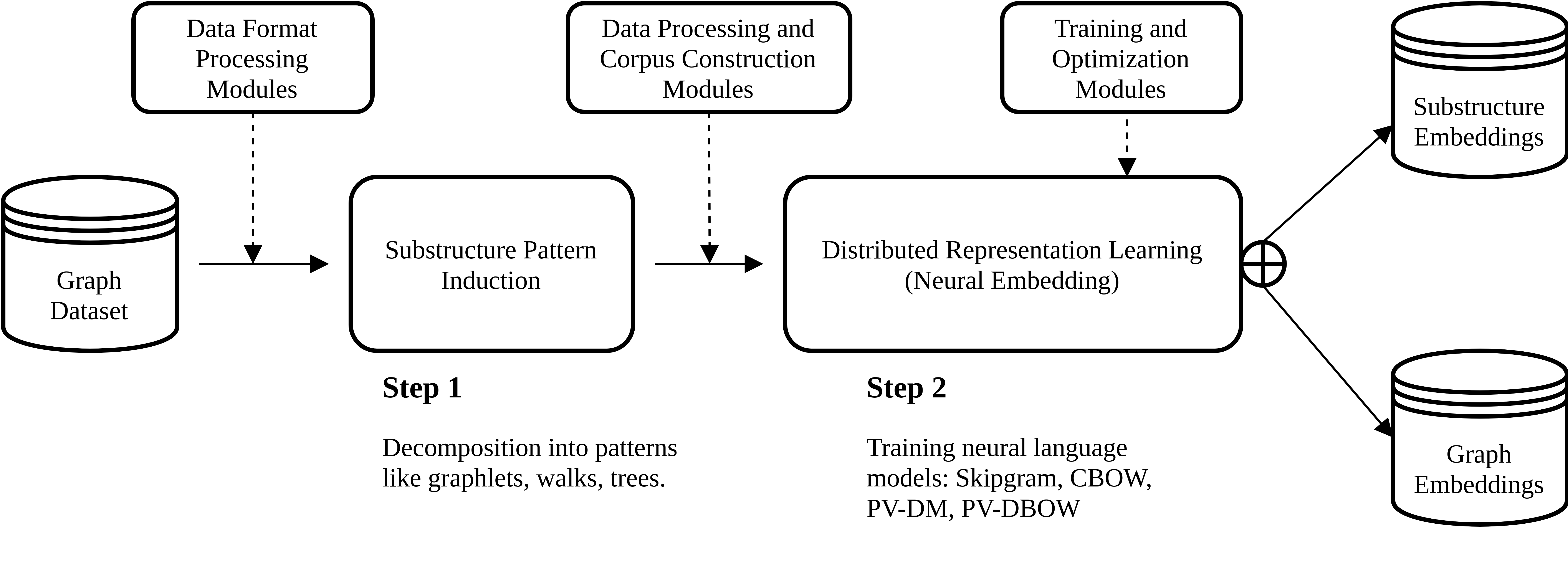}
    \caption{The two-stage design methodology for creating distributed representations of graphs and the various modules (in rectangles) included in Geo2DR to support this process. Each module can also be used independently for other tasks as mentioned in Section \ref{sec:overview} and \ref{sec:empiricaleval}.}
    \label{fig:geo2dr_example}
\end{figure*}

\section{Background and Related Work}

The approach towards distributive modelling of graphs was pioneered by Yanardag and Vishwanathan \yrcite{deepgraphkernels}. They observed that many graph kernel methods can be formulated as instances of the R-Convolutional framework. Herein, the similarity between different graphs is computed by decomposing graphs into discrete substructure patterns such as graphlets, shortest paths, and rooted subgraphs. This produces a $|\mathbf{V}|$-dimensional bag-of-words or pattern frequency vectors for each graph where $\mathbf{V}$ is the set of the unique patterns induced over all the graphs in a dataset. The graphs and their induced substructure patterns are input to a kernel function, such as counting the common substructures across pattern frequency vectors. This defines the relation or similarity measure between the graphs to construct the kernel matrix for use with kernel methods such as SVMs.

Yanardag and Vishwanathan \yrcite{deepgraphkernels} further observed that as the size of graphs and the specificity of substructure patterns to be induced from graphs increases (via lengthening walks/paths, increasing the number of nodes in graphlet patterns) graphs are represented by extremely high dimensional pattern frequency vectors. As a result, only few substructure patterns are common across any given set of graphs producing sparse solutions where each graph is more similar to itself, a phenomenon known as \textit{diagonal dominance}. To tackle this issue the authors proposed the use of neural language models which exploit the distributive hypothesis \cite{harris} to learn smooth low dimensional \textit{distributed representations} of the substructures and construct graph kernel matrices. This was quickly followed up by works such as the aptly named Graph2Vec \cite{graph2vec} and Anonymous Walk Embeddings \cite{anonymouswalkembeddings} (AWE). These proposed different substructure patterns graphs could be reduced to and the use of Doc2Vec variants \cite{doc2vec} to build distributed representations of whole graphs directly. A brief primer can be found in Appendix A.

Geo2DR provides various modules that can be used as ``building blocks" to rapidly construct systems capable of learning such distributed representations of both substructure patterns and whole graphs of arbitrary size. Existing libraries for GNNs \cite{relationalinductivebias,wang2019dgl,goyal,Fey2019} would require a substantial shift in philosophical focus from constructing message passing schemes and pooling methods. Hence Geo2DR is a complementary library alongside existing toolkits enabling researchers a broader range of options and tools for graph representation learning.

\section{Overview of Geo2DR}\label{sec:overview}
Geo2DR contains various "building blocks" for rapid construction of methods capable of learning distributed representations of graphs. The conceptual framework for unsupervised learning of the representations for substructures and entire graphs is based around a simple two stage design methodology summarised in Figure 1. 

    \textbf{Induction of descriptive substructure patterns}: The first step consists of inducing discrete substructure patterns such as graphlets, rooted subgraphs, or anonymous walks within and across the dataset of graphs to construct a shared vocabulary and \textit{corpus} dataset contextualizing the patterns and graphs. One may also use the output pattern distributions at this stage to construct a variety of graph kernels. 

    \textbf{Learning distributed vector representations}: The second stage consists of utilising the distributive hypothesis \cite{harris} to learn distributed representations of graphs contextualised by the induced substructure patterns. Embedding methods which exploit the distributive hypothesis such as skipgram (Mikolov et al., 2014) can be used to learn fixed-size vector embeddings of substructure patterns or whole graph in an unsupervised manner.

\begin{table*}
\small
\centering
\caption{Table characterising each of the existing published methods by the substructure patterns induced and associated embedding method to create the graph kernel matrix (for DGK models) or graph embeddings.}
\label{tab:existing_models}
\begin{tabular}{@{}llll@{}}
\toprule
Method & Induced substructure pattern & Embedding method & Object embedded \\ \midrule
DGK-WL & WL rooted subgraphs & Skipgram or CBOW & Substructure patterns \\
DGK-SP & Shortest paths & Skipgram or CBOW & Substructure patterns \\
DGK-GK & Graphlets & Skipgram or CBOW & Substructure patterns \\
Graph2Vec & WL rooted subgraphs & PV-DBOW & Whole graphs \\
AWE-DD & Anonymous walks & PV-DM & Whole graphs \\ \bottomrule
\end{tabular}%
\end{table*}

Combination of Geo2DR's modules for decomposition and distributed representation learning methods can be used to quickly replicate existing models such as those shown in Table \ref{tab:existing_models}. Consistent input/output interfaces were implemented across modules to encourage exploration of novel methods. For example, one could create a "novel" unpublished method combining existing modules on inducing shortest path patterns and learning graph-level embeddings with PV-DBOW. This sort of experimentation fosters understanding and better control of the inductive biases involved in a graph learning task. We hope it would also encourage the creation of custom modules that can plug and play with the rest of the framework to create truly novel methods.

Practically, the library is centered around three subpackages under Geo2DR. The \texttt{data} subpackage, contains modules for transforming data formats used by popular dataset repositories such as Kersting et al. \yrcite{graphkerneldatasite} into consistent formats used by the decomposition algorithms implemented in Geo2DR. In Geo2DR, we chose to use the GEXF (Graph Exchange XML Format) as permanent storage format for individual instances of the graphs. This is because the format is compatible with network analysis software such as Gephi and NetworkX for detailed inspection.

The modules within the \texttt{decomposition} subpackage contain algorithms for inducing the substructure patterns in the graphs and forming vocabularies. The outputs of these algorithms are directly compatible with our PyTorch implementations of neural language models to utilize GPUs as well as those in Gensim \cite{gensim}. This essentially describes the packages and modules necessary for Step 1 of the process. The final subpackage \texttt{embedding\_methods} contains modules for constructing corpus datasets and neural language models to build the distributed representation learning methods of Step 2. Several \texttt{Trainer} classes are also included which serve as battery-included corpus and neural net combinations that can be used to construct common architecture setups. 

Existing methods for learning distributed representations as in Table \ref{tab:existing_models} and several graph kernels can all be implemented using the modules and conceptual framework presented. We have included all methods as examples within the repository to get users started on creating their own variations. A brief code example is included in Appendix B.

\begin{table*}
\caption{Random-split 10 fold cross-validation performance of SVM using RBF kernel on bag-of-words vectors of normalised frequencies of induced substructure patterns. Best scores or those within error of best are bolded. OOM denotes out-of-memory.}
\label{tab:mlekernel}
\centering
\resizebox{0.9\textwidth}{!}{%
\begin{tabular}{@{}lllllll@{}}
\toprule
Substructure pattern & MUTAG & ENZYMES & PROTEINS & NCI1 & REDDIT-B & IMDB-M \\ \midrule
WL Rooted Subgraphs & \textbf{88.95 $\pm$ 7.96} & \textbf{56.33 $\pm$ 6.18} & 74.29 $\pm$ 2.55 & \textbf{83.94 $\pm$ 1.99} & 77.35 $\pm$ 4.35 & 48.60 $\pm$ 4.33 \\
Shortest Paths & 83.68 $\pm$ 7.24 & 41.67 $\pm$ 4.83 & \textbf{74.73 $\pm$ 2.04} & 70.95 $\pm$ 1.95 & OOM & \textbf{50.20 $\pm$ 3.84} \\
Graphlets & 83.16 $\pm$ 6.16 & 25.33 $\pm$ 3.48 & 70.36 $\pm$ 3.59  & 54.09 $\pm$ 7.61 & 78.25 $\pm$ 2.71 & 44.40 $\pm$ 4.17 \\
Anonymous Walks & 80.53 $\pm$ 6.68 & 27.33 $\pm$ 6.23 & 71.87 $\pm$ 2.05 & 66.08 $\pm$ 2.21 & \textbf{81.30 $\pm$ 2.49} & 38.20 $\pm$ 3.91 \\ \bottomrule
\end{tabular}%
}
\end{table*}

\begin{table*}
\caption{Graph classification performance over random-split 10 fold cross-validation in each of the re-implemented systems with standard deviation. Best scores or those within error of best are bolded. OOM denotes out-of-memory.}
\label{tab:deepmethods}
\centering
\resizebox{0.8\textwidth}{!}{%
\begin{tabular}{@{}lllllll@{}}
\toprule
Method & MUTAG & ENZYMES & PROTEINS & NCI1 & REDDIT-B & IMDB-M \\ \midrule
DGK-WL & \textbf{88.42 $\pm$ 8.42} & 41.00 $\pm$ 1.83 & 72.08 $\pm$ 0.74 & \textbf{77.54 $\pm$ 3.91} & OOM & 47.82 $\pm$ 0.79 \\
DGK-SP & 84.03 $\pm$ 7.16 & 44.27 $\pm$ 2.26 & \textbf{76.93 $\pm$ 2.56} & 69.22 $\pm$ 5.29 & OOM & \textbf{49.71 $\pm$ 1.18} \\
DGK-GK & 84.21 $\pm$ 6.74 & 23.61 $\pm$ 3.14 & 69.77 $\pm$ 3.13 & 53.92 $\pm$ 4.81 & 78.32 $\pm$ 1.92 & 44.40 $\pm$ 4.18 \\ \midrule
Graph2Vec & 84.91 $\pm$ 2.79 & \textbf{51.77 $\pm$ 1.75} & 74.05 $\pm$ 2.28 & 71.34 $\pm$ 2.12 & \textbf{81.25 $\pm$ 2.64} & 47.11 $\pm$ 1.42 \\
AWE-DD & 79.29 $\pm$ 2.92 & 23.76 $\pm$ 1.74 & 69.70 $\pm$ 1.29 & 63.54 $\pm$ 1.82 & \textbf{81.46 $\pm$ 1.75} & 40.53 $\pm$ 6.42 \\ \bottomrule
\end{tabular}%
}
\end{table*}

\section{Empirical Evaluation} \label{sec:empiricaleval}
As a form of validation for the various implemented modules, we empirically evaluate re-implementations of existing models using Geo2DR. Table \ref{tab:existing_models} describes the induced substructure pattern and neural language model driving each method. We performed a series of common benchmark graph classification tasks under homogeneous data and evaluation scenarios giving a fairer picture of how they compare.

All datasets were downloaded from the benchmark dataset repository by Kersting et al. \yrcite{graphkerneldatasite} and processed into the format used by Geo2DR with the included data formatter. In each of the datasets the discrete node labels are exposed, but not the edge labels. For unlabelled datasets such as REDDIT-B, the node was labelled by their degree following practice of Shervashidze et al. \yrcite{wlkernel} to enable methods such as the WL rooted subgraph decomposition to induce patterns in the graphs; this was also applied to methods which can directly handle unlabelled graphs for conformity. As these datasets are standard benchmarks we have left specific descriptive details in Appendix C. 

For all experiments, attempts were made to follow the hyperparameter setups described in the published papers of the original methods, with best-guess settings where details were not published. As we look at several kernels and embedding models specific hyperparameter ranges can be found in Appendix D. In all cases, the same off-the-shelf SVM implemented in SciKit-Learn \cite{scikit-learn} was used with an RBF kernel trick for the supervised classification task on the graph embeddings learned. $C$ values were estimated over the set (0.001, 0.01, 0.1, 1, 10, 100). We report the average score of 10 iterations of training and applying 10 fold cross-validation using the SVM over random data splits with individual training restarts in all cases. The exact setups of the experiments can be replicated using the experiment replication code provided within the Github repository\footnote{https://github.com/paulmorio/geo2dr/tree/master/replication}.

\textbf{Graph kernels:} We start with an experiment suite based on the substructure patterns alone, using the decomposition algorithms to construct normalised bag-of-words frequency vectors for each of the graphs. Table \ref{tab:mlekernel} records the mean and standard deviation of randomly split 10 fold cross-validation using the SVM described above. The results closely match that of the published methods in \cite{deepgraphkernels,wlkernel,shortestpathkernel,anonymouswalkembeddings}.

\textbf{Deep graph kernels and graph embeddings:} Most of our experiments in Table \ref{tab:deepmethods} show a high reproducibility of the results published by the original proposers. Some discrepancies are to be expected due to the homogenised data setup, unpublished hyperparameter settings, and standardised neural architectures, but best effort was made by consulting original source code and communications with the authors. In particular, for AWE-DD, we do not use edge-labels for homogeneity of the experiment evaluation whilst the original paper used them if they gave better scores.

\textbf{Runtime experiments and improvements in Geo2DR:} Table \ref{tab:runtimev2} contains the average total training times incurred over 100 epochs, performed ten times with one standard deviation on a single quad-core Intel i5-4690 CPU. Comparison is drawn between the original reference implementation made available by each of the original papers and its re-implemented counterpart in Geo2DR. All methods were trained and compared on the MUTAG dataset as this was the only common dataset included in the reference implementations. None of the original reference implementations have scripts or tools to transform the publicly available datasets they used into the proprietary formats used by their own implementations, making reproduction difficult. This is why we have included data processing tools directly into the Geo2DR library to address this common limitation for the future. 


\begin{table}
\centering
\caption{Total training run time (seconds) over 100 epochs on MUTAG. Bold text refers to lowest time taken for training or are within error bounds of being the fastest.}
\label{tab:runtimev2}
\resizebox{\columnwidth}{!}{%
\begin{tabular}{@{}llll@{}}
\toprule
Method & \begin{tabular}[c]{@{}l@{}}Original\\ reference \\ implementation\end{tabular} & \begin{tabular}[c]{@{}l@{}}Only Geo2DR\\ PyTorch modules\end{tabular} & \begin{tabular}[c]{@{}l@{}}Geo2DR with \\ compatible libraries \\ Gensim/TensorFlow  \end{tabular} \\ \midrule
DGK-WL & \textbf{3.06 $\pm$ 0.15} & 3.33 $\pm$ 0.07 & \textbf{3.19 $\pm$ 0.08} \\
DGK-SP &\textbf{6.95 $\pm$ 0.23} &\textbf{6.86 $\pm$ 0.27} & 7.39 $\pm$ 0.08 \\
DGK-GK & \textbf{9.46 $\pm$ 0.69} & 19.41 $\pm$ 0.49 & \textbf{9.89 $\pm$ 0.74} \\
Graph2Vec & \textbf{8.86 $\pm$ 0.05} & 10.64 $\pm$ 0.11 & \textbf{8.88 $\pm$ 0.06} \\
AWE-DD &  1231.75 $\pm$ 21.81 & \textbf{314.84 $\pm$ 8.91} & --- \\ \bottomrule
\end{tabular}%
}
\end{table}

\section{Conclusion} 

Through the characterisation of existing methods, and the reproduction of their results in Geo2DR, we have shown that the library is a successful amalgamation of the various components that enable learning distributed representations of graphs. Using the simple design methodology, one can quickly re-implement existing models, an increasingly important part of reproducible research and designing novel architectures. By exploiting the modular structure and compatibility with other software and libraries the set of tools for constructing learning methods is broadened without having to deal with different data formats, language paradigms and workflows used by individual implementations. Using a host of re-implemented methods also allows for more homogenised experiment suites that can be used to more fairly compare existing and new methods in future research efforts.  Geo2DR is available now with numerous examples and documentation as a starting point. The library will continue to evolve to add new components, compatibility with other libraries, tutorials, and accommodate new developments in the field.

\section*{Acknowledgements}
Foremost, we would like to thank Dr. Yanardag \cite{deepgraphkernels}, Dr. Narayanan \cite{graph2vec} and Dr. Ivanov \cite{anonymouswalkembeddings} for their correspondence, and making reference code available publicly. Furthermore we would like to thank the members of the AI Group at the Computer Laboratory for their patience and numerous proof readings of this work.

\bibliography{example_paper}
\bibliographystyle{icml2020}

\appendix

\section{Brief Primer on Learning Distributed Representations of Graphs}

\begin{figure*}
    \centering
    \includegraphics[width=0.9\textwidth]{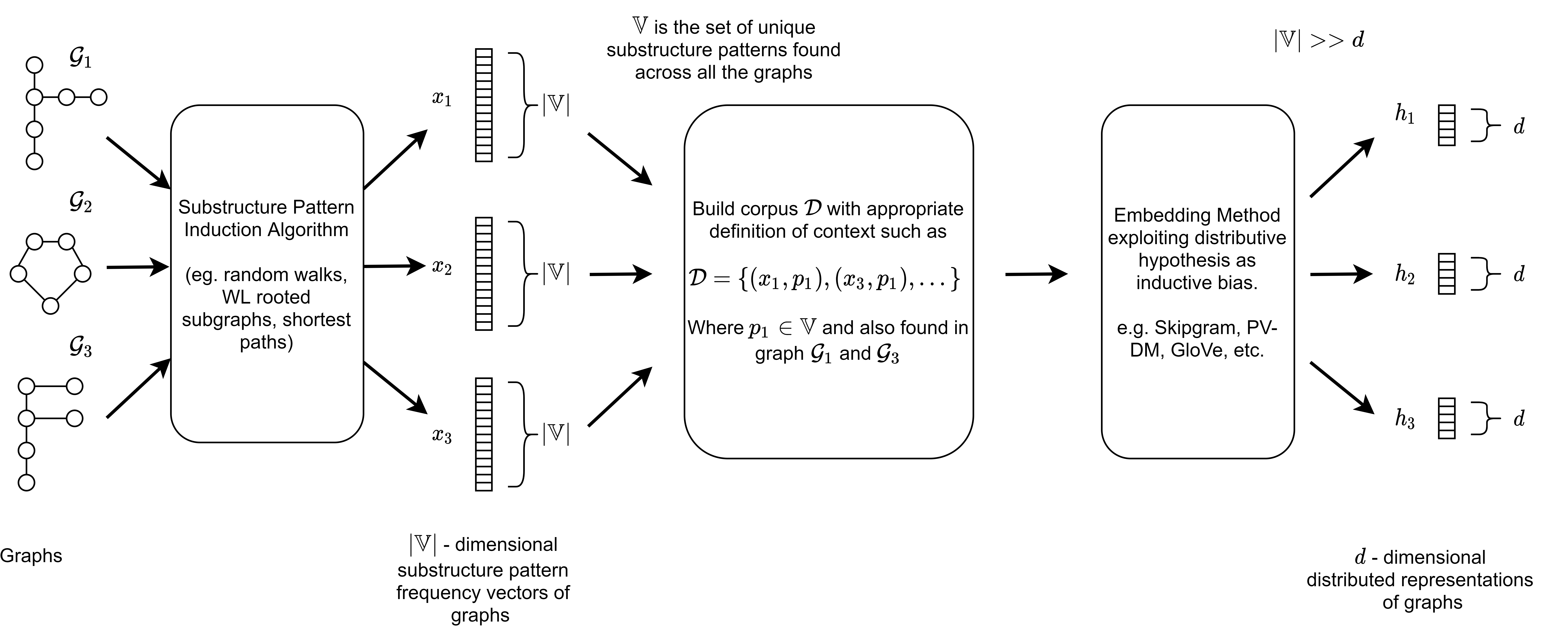}
    \caption{A conceptual framework for how methods for learning distributed representations of graphs are constructed, which guides the method design principles in Geo2DR.}
    \label{fig:drg}
\end{figure*}

Here we provide a brief and simplified primer on learning distributed representations of graphs. This will not fully describe the various intricacies of existing methods, but cover a conceptual framework common to almost all distributed representations of graphs particularly for learning representations of substructure patterns and whole graphs. Figure \ref{fig:drg} is a diagrammatic representations of this conceptual framework.

Given a set of graphs $\mathbb{G} = \{ \mathcal{G}_1, \mathcal{G}_2, ... \mathcal{G}_n \}$ one can induce discrete substructure patterns such as shortest paths, rooted subgraphs, graphlets, etc. using side-effects of algorithms such as the Floyd-Warshall or Weisfeiler-Lehman Graph Isomorphism test, and so on. This can be used to produce pattern frequency vectors $\mathbf{X} = \{\mathbf{x}_1, \mathbf{x}_2, ..., \mathbf{x}_n \}$ describing the occurrence frequency of substructure patterns over a shared vocabulary $\mathbb{V}$. $\mathbb{V}$ is the set of unique substructure patterns induced across all of the graphs in the dataset $\mathbb{G}$. 

Classically one may directly use these pattern frequency vectors within standard machine learning methods using vector inputs to perform some task. This is the approach taken by a variety of graph kernels \cite{deepgraphkernels, vishreview}. Unvfortunately, as the graphs of $\mathbb{G}$ and subtructure patterns induced become more complex through size or specificity, the number of induced patterns increases dramatically. This, in turn, causes the pattern frequency vectors of $\mathbf{X}$ to be extremely sparse and high-dimensional. The high specificity of the patterns and the sparsity of the pattern frequency vectors cause a phenomenon known as diagonal dominance across the kernel matrices wherein each graph becomes more similar to itself and dissimilar from others, degrading the classification performance \cite{deepgraphkernels}.

To address this issue it is possible to learn dense and low dimensional distributed representations of graphs that are inductively biased to be similar when they contain similar substructure patterns and dissimilar when they do not. To achieve this, the construction of a corpus dataset $\mathcal{D}$ is required detailing the target-context relationship between a graph and its induced substructure as in our example or a substructure pattern to other substructure patterns. In the simplest form for graph-level representation learning one can implement $\mathcal{D}$ as tuples of graphs and substructure pattern $(\mathcal{G}_i, p_j) \in \mathcal{D}$ if $p_j \in \mathbb{V}$ and $p_j \in \mathcal{G}_i$. 

The corpus is utilised with a method that incorporates Harris' distributive hypothesis \yrcite{harris} to learn the distributed representations of graphs. skipgram, cbow, PV-DM, PV-DBOW \cite{word2vec, doc2vec} are a few examples of neural embedding methods that incorporate this inductive bias and are all present in the Geo2DR library. In skipgram with negative sampling, as used in Graph2Vec \cite{graph2vec}, the distributed representations can be learned by optimizing

\begin{align*}
\mathcal{L} = \sum_{\mathcal{G}_i \in \mathbb{G}} \sum_{p \in \mathbb{V}} |\{{(\mathcal{G}_i, p ) \in \mathcal{D}}\}| (\log \sigma(\Phi_i \cdot \mathcal{S}_{p}) \\
+ k \cdot \mathbb{E}_{p_N \in P_D}[\log \sigma(-\Phi_i \cdot p_N)] 
\end{align*}

over the corpus observations \noindent where $\Phi \in \mathbb{R}^{|\mathbb{G}| \times d}$ is the $d$ dimensional matrix of graph embeddings we desire of the graph dataset $\mathbb{G}$, and $\Phi_i$ is embedding for $\mathcal{G}_i \in \mathbb{G}$. Similarly, $\mathcal{S} \in \mathbb{R}^{|\mathbb{V}| \times d}$ are the $d$ dimensional embeddings of the substructure patterns in the vocabulary $\mathbb{V}$ so $\mathcal{S}_p$ represents the vector embedding corresponding to substructure pattern $p$. The embeddings of the substructure patterns are also tuned but ultimately not used, as we are interested in the graph embeddings in $\Phi$. $k$ is the number of negative samples with $t_N$ being the sampled context pattern, drawn according to the empirical unigram distribution $P_D (p) = \frac{|\{p | \forall G_i \in \mathbb{G}, (G_i, p) \in \mathcal{D}\}|}{|D|}$.

The optimization of the above utility function creates the desired distributed representations of the targets in $\Phi$, in this the case graph-level embeddings. These may be used as input for any downstream machine learning task and method that take vector inputs. The distributed representations benefit from having lower dimensionality than the pattern frequency vectors, in other words $|\mathbb{V}| >> d$, being non-sparse, and being inductively biased via the distributive hypothesis in an unsupervised manner. For more in-depth reading we recommend \cite{harris, word2vec, doc2vec, deepgraphkernels, graph2vec}.

\section{Code Example}

We present a construction of a simplified Graph2Vec model and training it to produce 32 dimensional distributed vector embeddings of the MUTAG graphs using Geo2DR modules. To start we need to download a dataset to study. We will use the well known MUTAG \cite{mutag} downloaded from the TU Dortmund Graph Kernel Benchmark website \cite{graphkerneldatasite}. Assume we have unpacked and saved the data into a directory called \texttt{org\_data/} so the dataset as downloaded will be within the directory as \texttt{org\_data/MUTAG/}.

Geo2DR uses the GEXF (Graph Exchange XML Format) as the permanent storage format for the graphs in a dataset. This is because it is compatible with network analysis software such as Gephi and NetworkX, and it is often useful to be able to study each graph individually; identified by a single file. Due to this design choice we need to transform the format of the downloaded dataset using tools available within the \texttt{data} subpackage as in the code sample below.

\begin{lstlisting}[language=Python, caption=Formatting the downloaded dataset into GEXF format]
from geometric2dr.data import DortmundGexf

gexifier = DortmundGexf("MUTAG","org_data/","data/")
gexifier.format_dataset()
\end{lstlisting}

\noindent This will result in the following dataset format:

\begin{itemize}
    \item \texttt{data/MUTAG/} : a directory containing individual \texttt{.gexf} files of each graph. A graph will be denoted by the graph IDs used in the original data. In this case graph 0 would be \texttt{data/MUTAG/0.gexf}
    \item \texttt{data/MUTAG.Labels} : a plain-text file with each line containing a graph’s file path and its classification label.
\end{itemize}

Given the preprocessed data we can now induce substructure patterns across the graph files. Here we will induce rooted subgraphs up to depth 2 using the Weisfeiler-Lehman node relabeling algorithm outlined in Shervashidze et al. \yrcite{wlkernel}.

\begin{lstlisting}[language=Python, caption= Inducing rooted subgraphs across the graphs of the dataset]
from geometric2dr.decomposition.weisfeiler_lehman_patterns import wl_corpus
import geometric2dr.embedding_methods.utils as utils

dataset_path = "data/MUTAG"
graph_files = utils.get_files(dataset_path,".gexf")

wl_depth = 2
wl_corpus(graph_files,wl_depth)
\end{lstlisting}

The \texttt{wl\_corpus()} function induces rooted subgraph patterns across the list of \texttt{.gexf} files in \texttt{graph\_files}, and builds a document for each graph describing the induced patterns within. These documents will have a special extension specific to each decomposition algorithm or can be set by the user. In this example the extension will be \texttt{.d2wl} to denote a Weisfeiler-Lehman decomposition to depth 2. Generating permanent files as a side effect of the graph decomposition process is useful for later study and also if we want to use the same induced patterns in the upcoming step of learning distributed representations of the graphs.

To learn distributed representations we need to construct a new target-context dataset. In Graph2Vec a graph is contextualised by the substructure patterns within it, and uses the PV-DBOW architecture with negative sampling to directly learn graph-level embeddings. Hence we use the \texttt{PVDBOWInMemoryCorpus} which is a extension of a standard \texttt{torch.utils.data.dataset} class. This can interface with a standard PyTorch dataloader to load the data into a \texttt{embedding\_methods.skipgram} class that we train in a loop using a simple and recognizable \texttt{torch.nn} workflow.

\begin{lstlisting}[language=Python, caption=Creating a target-context dataset then attaching a dataloader that feeds the corpus data into a skipgram model and training it.]
import torch
import torch.optim as optim
from torch.utils.data import DataLoader
from geometric2dr.embedding_methods.pvdbow_data_reader import PVDBOWInMemoryCorpus
from geometric2dr.embedding_methods.skipgram import Skipgram

# Instantiate corpus dataset, dataloader and skipgram 
# architecture
corpus = PVDBOWCorpus(dataset_path,".d2wl")
dataloader = DataLoader(corpus,batch_size=1000,shuffle=False,collate_fn=corpus.collate)
skipgram = Skipgram(num_targets=corpus.num_graphs,vocab_size=corpus.num_subgraphs,emb_dimension=32)

optimizer = optim.SGD(skipgram.parameters(), lr=0.1)
for epoch in range(100):
  for i, sample_batched in enumerate(dataloader):
    if len(sample_batched[0]) > 1:
      pos_target = sample_batched[0]
      pos_context = sample_batched[1]
      neg_context = sample_batched[2]
      
      optimizer.zero_grad()
      loss = skipgram.forward(pos_target,pos_context,neg_context)
      loss.backward()
      optimizer.step()

final_graph_embeddings = skipgram.target_embeddings.weight
\end{lstlisting}

The completion of the training provides the final graph embeddings. As this is such a common proces, Geo2DR also comes with a number of \texttt{Trainer} classes which build corpus datasets, loaders, train neural language models, and save their outputs. All of the code above can be replaced with this short trainer.

\begin{lstlisting}[language=Python, caption= Trainer example of performing all of listing 1.3]
from geometric2dr.embedding_methods.pvdbow_trainer import InMemoryTrainer

trainer = InMemoryTrainer(corpus_dir=dataset_path,extension=".d2wl", output_fh="graph_embeddings.json",emb_dimension=32,batch_size=1000,epochs=100,initial_lr=0.1,min_count=0)
trainer.train()
final_graph_embeddings = trainer.skipgram.give_target_embeddings()
\end{lstlisting}

\section{Supplementary: Dataset Details}

\begin{table*}\label{table:datasets}
\centering
\caption{Descriptive information about datasets used in the experimental evaluation. N refers the number of graphs in the datasets. C is the number of graph classification labels. Avg. Nodes and Avg. Edges denote the average number of nodes and edges found in the graphs of the dataset respectively. Finally Node Labels indicates whether the nodes are discretely labelled. The * refers to datasets which originally do not have node labels, but are subsequently labelled by their degree as described in Shervashidze et al. \yrcite{wlkernel}}
\label{tab:datasets}
\resizebox{0.75\textwidth}{!}{
\begin{tabular}{@{}llllll@{}}
\toprule
Dataset  & N    & C & Avg. Nodes & Avg. Edges & Node Labels \\ \midrule
MUTAG \cite{mutag}    & 188  & 2 & 17.93      & 19.79      & Yes         \\
ENZYMES \cite{proteins}  & 600  & 6 & 32.63      & 62.14      & Yes         \\
PROTEINS \cite{proteins} & 1113 & 2 & 39.06      & 72.82      & Yes         \\
NCI1 \cite{nci}    & 4110 & 2 & 29.87      & 32.3       & Yes         \\
REDDIT-B \cite{deepgraphkernels} & 2000 & 2 & 429.63     & 497.75     & No*         \\
IMDB-M \cite{deepgraphkernels}  & 1500 & 3 & 13         & 65.94      & No*         \\ \bottomrule
\end{tabular}%
}
\end{table*}

Table 5 contains descriptive information about each of the datasets as they were used within the empirical evaluation described in Section \ref{sec:empiricaleval} of the main paper. All of the datasets are commonly used benchmark datasets downloaded from Kersting et al.'s \yrcite{graphkerneldatasite} repository\footnote{ls11-www.cs.tu-dortmund.de/staff/morris/graphkerneldatasets}. After downloading the datasets they were processed into the format used by Geo2DR with the included data formatter. In each of the datasets the discrete node labels are exposed, but not the edge labels. For unlabelled datasets such as REDDIT-B and IMDB-M, the nodes are labelled by their degree as in Shervashidze et al. \yrcite{wlkernel} to enable methods such as the WL rooted subgraph decomposition to induce patterns in the graphs. This was also applied to methods which can directly handle unlabelled graphs for conformity.

The graphs come from a variety of contexts and domains. MUTAG, ENZYMES and PROTEINS are datasets which have their roots in bioinformatics research. The graphs within them represent molecules with nodes representing atoms and edges denoting chemical bonds or spatial proximity between different atoms. Graph labels describe different properties of the molecules such as mutagenicity or whether a protein is an enzyme. NCI1 is a chemoinformatics dataset describing compounds screened for their ability to surpress or inhibit the growth of a panel of human tumor cell lines. REDDIT-B and IMDB-M are social network based datasets. In REDDIT-B each graph corresponds to an online discussion thread where nodes correspond to users, and there is an edge between the nodes if at least one responded to another's comment. IMDB-M is a movie collaboration dataset where each graph corresponds to an ego-network of an actor or actress.

\section{Supplementary: Hyperparameter Selections of Re-implemented Methods}
For each of the methods described in Section \ref{sec:empiricaleval} we prescribed a grid search over the following hyper-parameter settings inspired by the settings of the original papers:

\subsection{Graph Kernels}
\begin{itemize}
    \item \textbf{WL Rooted Subgraphs:} Rooted subgraphs up to depth 2 induced.
    \item \textbf{Shortest Paths:} Shortest paths of all pairs of nodes induced.
    \item \textbf{Graphlets:} Graphlets of size 7 induced, sampling 100 graphlets per graph.
    \item \textbf{Anonymous Walks:} Anonymous walks of length 10 induced exhaustively from each node in the graph.
\end{itemize}

\subsection{Deep Graph Kernels and Graph Embeddings}
\begin{itemize}
    \item \textbf{DGK-WL:} Rooted subgraphs of up to depth 2 induced. Trained Skipgram model with negative sampling using 10 negative samples with an Adam optimiser for 5 and 100 epochs using batch sizes of 10000 and 1000 with an initial learning rate of 0.1 and 0.01 adjusted by a cosine annealing scheme. Substructure embedding sizes of 2, 5, 10, 25, 50 dimensions were generated. Graph kernels were constructed using the formulation described in Yanardag and Vishwanathan \yrcite{deepgraphkernels}.
    \item \textbf{DGK-SP:} Shortest paths of all pairs of nodes induced. Trained Skipgram model with negative sampling using 10 negative samples with an Adam optimiser for 5 and 100 epochs using batch sizes of 10000 and 1000 with an initial learning rate of 0.1 and 0.01 adjusted by a cosine annealing scheme. Substructure embedding sizes of 2, 5, 10, 25, 50 dimensions were generated. Graph kernels were constructed using the formulation described in Yanardag and Vishwanathan \yrcite{deepgraphkernels}.
    \item \textbf{DGK-GK:} Graphlets of size 7 induced, sampling 2, 5, 10, 25, and 50 graphlets for each graph. Trained Skipgram model with negative sampling using 10 negative samples with an Adam optimiser for 5 and 100 epochs using batch sizes of 10000 and 1000 with an initial learning rate of 0.1 and 0.01 adjusted by a cosine annealing scheme. Substructure embeddings of 2, 5, 10, 25, 50 dimensions were generated. Graph kernels were constructed using the formulation described in Yanardag and Vishwanathan \yrcite{deepgraphkernels}.
    \item \textbf{Graph2Vec:} Rooted subgraphs of up to depth 2 induced. Trained over PV-DBOW (Skipgram) model with negative sampling using 10 negative samples with an Adam optimiser for 25, 50, 100 epochs and batch sizes of 512, 1024, 2048, 10000 with an initial learning rate of 0.1 adjusted by a cosine annealing scheme. Graph embeddings of 128 and 1024 dimensions were learned.
    \item \textbf{AWE-DD:} Anonymous walks of length 10 induced exhaustively. Trained over PV-DM architecture with negative sampling using 10 negative samples with an Adagrad optimiser (as in reference implementation) for 100 epochs with batch sizes 100, 500, 1000, 5000, 10000 with an initial learning rate of 0.1. Window-sizes of 4, 8, 16 were used to extract context anonymous walks around the target anonymous walk in the PV-DM architecture.
\end{itemize}

\end{document}